\documentclass[a4paper,12pt]{article}
\usepackage[utf8]{inputenc}
\usepackage[english]{babel}
\usepackage{authblk}
\usepackage{graphicx}
\usepackage{mathptmx}
\usepackage[singlespacing]{setspace}
\usepackage[headheight=1in,margin=1in]{geometry}
\usepackage{fancyhdr}
\usepackage{comment}
\usepackage{wrapfig}
\usepackage[font=footnotesize]{caption}

\usepackage{xcolor}
\usepackage{amsfonts}
\usepackage{amsmath}
\usepackage{csquotes}
\usepackage{wrapfig}
\usepackage{subcaption}
\usepackage{url}
\usepackage{hyperref}

\usepackage{natbib}

\title{Automatic Detection of Research Values from Scientific Abstracts Across Computer Science Subfields}

\author[1]{Hang Jiang\footnote{Work done during HJ's internship at Allen Institute for AI. This paper was initially accepted and presented at the 3rd International Conference of Science of Science \& Innovation (2024) as an extended abstract.}}
\author[2]{Tal August}
\author[2]{Luca Soldaini}
\author[2]{Kyle Lo}
\author[2]{Maria Antoniak}
\affil[1]{Massachusetts Institute of Technology}
\affil[2]{Allen Institute for AI}

\date{}

\begin{document}

% \vspace{-2em}
\maketitle
\thispagestyle{fancy}

\vspace{-2em}

% \section*{Extended Abstract}

\section{Background}

The field of computer science (CS) has rapidly evolved over the past few decades, providing computational tools and methodologies to different fields and forming new interdisciplinary communities. This growth in CS has significantly impacted institutional practices and relevant research communities. Therefore, it is crucial to explore what specific \textbf{research values}, known as \textbf{basic and fundamental beliefs that guide or motivate research attitudes or actions}, CS-related research communities promote. We are inspired by prior research that has manually analyzed research values from a small sample of machine learning papers \citep{facct}. We extend that study by annotating additional data and testing automatic methods for value detection, as no prior work has studied the automatic detection of research values from large-scale scientific texts across different research subfields. This paper introduces a detailed annotation scheme featuring \textbf{ten research values} that guide CS-related research\footnote{Data and code are available at: \url{https://github.com/hjian42/CSValues}}. Based on the scheme, we build value classifiers to scale up the analysis and present an analysis of 226,600 paper abstracts from 32 CS-related subfields and 86 popular publishing venues over ten years. 

% explored the automatic detection of human values from arguments \cite{kiesel2022identifying} and 

\section{Research Value Scheme} 

Building on \citet{facct}, we define that a sentence encodes research value(s) if there is any expression of why something is desirable or undesirable. Through qualitative coding, three authors of the paper manually identified sentences discussing research values from 100 sampled abstracts and arrived at ten popular research values across different CS-related subfields. These values include (1) Performance, (2) Novelty, (3) Efficiency, (4) Generalizability, (5) Openness, Reproducibility, Collaboration, \& Future Work (Openness), (6) Simplicity, (7) Phenomenon Understanding \& Theoretical Grounding (Understanding), (8) Fairness, Bias, Privacy \& Ethics (Integrity), (9) Societal Implications (Society), (10) Usability. Each research value is defined in a codebook with representative examples provided.

\section{Data Sampling \& Annotation} 

We first curate a list of 86 popular CS-related research venues by consulting \url{csrankings.org} and sending questionnaires to 17 doctoral researchers from different communities. We then downloaded 226,600 papers from 2013 to 2022 with the internal PyS2 library. We randomly sampled 12 abstracts from each venue and randomly picked one sentence from each abstract for the annotation task, resulting in 1,032 sentences for annotation. Two authors of the paper double-annotated these sentences and held rounds of discussions to determine a single ``consensus label'' for whether each sentence contains a given research value. Due to the challenging nature of the value detection task, two authors annotate one value at a time. We split the annotated data into train (40\%), validation (30\%), and test (30\%) sets.

\section{Task Formulation \& Method} 

We formulate the research value detection task as follows: given a sentence $s$ from an abstract, our target is to identify if a specific value $v_i$ from a set of values $V$ is encoded in the sentence. We build lexicon-based value classifiers and report their performance on the test set. We find that lexicon-based methods achieve decent accuracy and F1 scores, with F1 ranging from 0.71 to 0.88 across ten research values. We also experiment with LLM prompting techniques \citep{ziems2024can} for value detection. Surprisingly, we find that lexicon-based methods outperform LLM-based few-shot prompting methods on at least five values. LLM-based methods slightly perform better in Novelty and Simplicity. We therefore rely on the lexicon-based methods for our subsequent analysis.
% and also hope to fine-tune small RoBERTa in the future.

\section{Analysis \& Findings} 

By running value classifiers on all the abstracts, we are able to characterize individual research communities by their most frequent and most distinctive values. 

First, we find that researchers in different CS subfields tend to emphasize different research values in abstracts. Traditional fields such as Computer Architecture and Operating Systems are known to emphasize Efficiency and Performance. On the other hand, AI-related subfields such as Computer Vision highlight Performance, Generalizability, and Novelty in their abstracts. Furthermore, we have noted that interdisciplinary CS+X fields such as Computational Biology \& Bioinformatics, Ethics of Computer Science, and Human-Computer Interaction (HCI), compared to traditional CS domains, are more likely to mention Openness and less likely to discuss Simplicity, Efficiency, and Novelty. The variation in these values demonstrates the diversity within CS communities in terms of their research practices. 

Secondly, we found that specific values are experiencing increased emphasis in abstracts over time. Between 2013 and 2022, an uptick in the discussion around seven values was observed in the abstracts across AI-related subfields, which include Robotics, Computer Vision, Natural Language Processing, and Machine Learning. This implies that AI-related researchers have adapted their practices to more clearly articulate the research values within their abstracts over time. However, this trend isn't consistent across all CS subfields. Intriguingly, interdisciplinary CS+X subfields have shown an increasing emphasis on values such as Openness, Performance, and Understanding in their abstracts over time. On the other hand, subfields in Computer Theory have demonstrated a decreasing pattern in discussions about Simplicity over time, suggesting that theories may have become more advanced and complex. Areas connected with traditional CS Systems have not shown any notable trends with regards to research values over time, showing that there is an established writing practice in these subfields. 

Third, we examine the linguistic patterns used to communicate research values in abstracts. We find that these values are occasionally expressed directly with phrases like ``novel solution'', ``simple framework'', or ``generalizable method''. However, more often they are subtly represented through conventional linguistic structures such as ``propose a technique'' or ``create a dataset'' to imply novelty, and ``works in a variety of tasks'' or ``adapts to unseen tasks'' to suggest generalizability. 

At last, we study the pairwise co-occurrence patterns of values through pointwise mutual information (PMI) and find that Openness and Simplicity tend to co-occur most frequently with the other five values. In addition, Novelty and Efficiency tend to co-occur with each other in abstracts. 

\section{Conclusion \& Future Directions} 

Our paper presents a detailed value scheme and conducts a study to uncover research values from scientific abstracts across different CS subfields over time. In the future, we plan to expand the study by (1) studying if these values co-occur and are correlated with citations, funding sources, author seniority, and co-authoring patterns, (2) extending the current analysis beyond abstracts such as introduction and even full papers, (3) extending the value scheme to non-CS fields to understand how values are expressed across disciplines.

% \begin{figure*}[t]
% 	\centering
% 	\includegraphics[width=1.0\textwidth]{figure}
% 	\caption{Ut sed risus ut ante consectetur ultricies}
% 	\label{figure}
% \end{figure*}

% \vspace{-4mm}
\small
\renewcommand{\markboth}[2]{}
\bibliographystyle{apalike}
\bibliography{bibliography.bib}

% \newpage
\section*{Datasheet Documentation}

This following section documents how we created the \texttt{ResearchValues} dataset following the \textit{datasheets for datasets} guideline \citep{datasheet}.

\bigskip
\noindent
1. \textbf{For what purpose was the dataset created? Was there a specific task in mind? Was there a specific gap that needed to be filled? Please provide a description.} The purpose of this dataset and study was to explore and analyze the research values promoted across different computer science (CS) subfields. We developed a detailed annotation scheme for 10 key CS research values and analyzed the distribution of these values in a large corpus of 226,600 CS paper abstracts. The goal was to characterize the diversity of values emphasized in different CS communities and track how they have evolved over time, providing insights into the research practices and priorities within the field. This work was inspired by a previous study by \citet{facct} which performed a similar analysis using hand-annotation and a smaller set of papers.

\bigskip
\noindent
2. \textbf{Who funded the creation of the dataset? If there is an associated grant, please provide the name of the grantor and the grant name and number.} The Semantic Scholar group at Allen Institute for AI (Ai2) funded this dataset.

\bigskip
\noindent
3. \textbf{What do the instances that comprise the dataset represent (for example, documents, photos, people, countries)? Are there multiple types of instances (for example, movies, users, and ratings; people and interactions between them; nodes and edges)? Please provide a description.} Each instance represents a sentence sampled from scientific paper abstracts in the computer science literature, and has been manually annotated by us with the research value(s) it expresses, from a set of 10 predefined research values.

\bigskip
\noindent
4. \textbf{How many instances are there in total (of each type, if appropriate)?} 1,032 instances in total.

\bigskip
\noindent
5. \textbf{Does the dataset contain all possible instances or is it a sample (not necessarily random) of instances from a larger set? If the dataset is a sample, then what is the larger set? Is the sample representative of the larger set (for example, geographic coverage)? If so, please describe how this representativeness was validated/verified. If it is not representative of the larger set, please describe why not (for example, to cover a more diverse range of instances, because instances were withheld or unavailable).} Each instance represents a sentence sampled from the larger corpus of 226,600 scientific paper abstracts in the computer science literature, spanning 32 CS-related subfields and 86 publishing venues over a 10-year period. From this larger corpus, we randomly sampled 12 abstracts from each venue and then randomly selected one sentence from each abstract, resulting in a sample of 1,032 annotated sentences. This sampling approach was used to create a small, representative dataset for the manual annotation of research values, though the paper does not provide explicit validation of how representative the sample is of the full corpus of CS literature.

\bigskip
\noindent
6. \textbf{What data does each instance consist of? ``Raw'' data (for example, unprocessed text or images) or features? In either case, please provide a description.} Each instance contains both the raw textual data (the sentence) as well as the research value labels annotated by us.

\bigskip
\noindent
7. \textbf{Is there a label or target associated with each instance? If so, please provide a description.} Each instance in the dataset has 10 predefined research values that were manually labeled for each sentence instance, with a binary label of 0 (not encoded) or 1 (encoded) for each value. These annotated research value labels, with the binary encoding, serve as the ground truth target information for the instances in this dataset. Users can fetch the raw abstract with the paper ID from the original dataset with Semantic Scholar API. 

\bigskip
\noindent
8. \textbf{Is any information missing from individual instances? If so, please provide a description, explaining why this information is missing (for example, because it was unavailable). This does not include intentionally removed information, but might include, for example, redacted text.} No, each instance contains the full information. 

\bigskip
\noindent
9. \textbf{Are relationships between individual instances made explicit (for example, users' movie ratings, social network links)? If so, please describe how these relationships are made explicit.} The dataset includes the paper ID for the paper from which each sentence was drawn.

\bigskip
\noindent
10. \textbf{Are there recommended data splits (for example, training, development/validation, testing)? If so, please provide a description of these splits, explaining the rationale behind them.} We divided the dataset of 1,032 annotated sentences into training (40\%), validation (30\%), and test (30\%) sets. We adopted these splits to ensure that each split has enough instances for model training or evaluation. 

\bigskip
\noindent
11. \textbf{Are there any errors, sources of noise, or redundancies in the dataset? If so, please provide a description.} Noise in the dataset stems from several sources. First, the use of automatic sentence tokenizers to split abstracts introduced imperfect segmentation, potentially affecting data quality and subsequent analysis. While imperfectly tokenized sentences remain in our dataset, we acknowledge this limitation. Second, gaps in the Semantic Scholar platform's coverage across years and venues result in incomplete sampling across computer science subfields. Finally, research value annotations reflect the authors' subjective assessments, introducing potential bias based on their experience and expertise.

\bigskip
\noindent
12. \textbf{Is the dataset self-contained, or does it link to or otherwise rely on external resources (for example, websites, tweets, other datasets)? If it links to or relies on external resources, a) are there guarantees that they will exist, and remain constant, over time; b) are there official archival versions of the complete dataset (that is, including the external resources as they existed at the time the dataset was created); c) are there any restrictions (for example, licenses, fees) associated with any of the external resources that might apply to a dataset consumer? Please provide descriptions of all external resources and any restrictions associated with them, as well as links or other access points, as appropriate.} The dataset is self-contained.

\bigskip
\noindent
13. \textbf{Does the dataset contain data that might be considered confidential (for example, data that is protected by legal privilege or by doctor-patient confidentiality, data that includes the content of individuals' non-public communications)? If so, please provide a description.} No, the dataset is fully public.

\bigskip
\noindent
14. \textbf{Does the dataset contain data that, if viewed directly, might be offensive, insulting, threatening, or might otherwise cause anxiety? If so, please describe why.} No.

\bigskip
\noindent
15. \textbf{How was the data associated with each instance acquired? Was the data directly observable (for example, raw text, movie ratings), reported by subjects (for example, survey responses), or indirectly inferred/derived from other data (for example, part-of-speech tags, model-based guesses for age or language)? If the data was reported by subjects or indirectly inferred/derived from other data, was the data validated/verified? If so, please describe how.} The dataset consists of directly observable raw text from computer science research paper abstracts. The labels of research values associated with each instance were were derived through human annotation with a validation process involving two annotators discussing and reaching agreement on the labels.

\bigskip
\noindent
16. \textbf{What mechanisms or procedures were used to collect the data (for example, hardware apparatuses or sensors, manual human curation, software programs, software APIs)? How were these mechanisms or procedures validated?} The data collection involved both software tools and manual human curation. We first consulted \url{csrankings.org} and conducted questionnaires using Google Forms with 17 doctoral researchers to curate a list of 86 CS-related venues. We then downloaded 226,600 papers from 2013 to 2022 from Semantic Scholar. After collecting the papers, we employed the pySBD sentence segmenter to split abstracts into sentences, followed by random sampling to select 12 abstracts per venue and one sentence per abstract. The final annotation process was done through manual human curation using the Prodigy software, where two authors double-annotated the sentences and held discussions to determine the final research values.

\bigskip
\noindent
17. \textbf{If the dataset is a sample from a larger set, what was the sampling strategy (for example, deterministic, probabilistic with specific sampling probabilities)?
} The dataset was sampled using a stratified, probabilistic sampling strategy from a larger set of 226,600 papers from 86 computer science venues over ten years (2013-2022). We used a two-stage sampling process: first, for each venue, we attempted to sample one paper from each year when possible, plus two additional abstracts from random years, totaling 12 abstracts per venue. Then, we randomly picked one sentence from each of these abstracts, resulting in 1,032 sentences for annotation. The sampling method, which drew data from different years, helped ensure the dataset represented trends over time.

\bigskip
\noindent
18. \textbf{Who was involved in the data collection process (for example, students, crowdworkers, contractors) and how were they compensated (for example, how much were crowdworkers paid)?} All authors of the paper participated in the initial qualitative coding process to identify potential research values. After establishing ten distinct research values, the first and last authors conducted double-annotation of the dataset to assign value labels.

\bigskip
\noindent
19. \textbf{Over what timeframe was the data collected? Does this timeframe match the creation timeframe of the data associated with the instances (for example, recent crawl of old news articles)? If not, please describe the timeframe in which the data associated with the instances was created.} The paper abstracts were downloaded on July 23, 2023. The labeling process for research values occurred between July 23, 2023 and November 30, 2024.

\bigskip
\noindent
20. \textbf{Were any ethical review processes conducted (for example, by an institutional review board)? If so, please provide a description of these review processes, including the outcomes, as well as a link or other access point to any supporting documentation.} The research was determined to be IRB exempt by Ai2 since it analyzes publicly available research paper abstracts and does not involve human subjects research beyond surveying doctoral researchers about CS venues.

\bigskip
\noindent
21. \textbf{Was any preprocessing/cleaning/labeling of the data done (for example, discretization or bucketing, tokenization, part-of-speech tagging, SIFT feature extraction, removal of instances, processing of missing values)? If so, please provide a description. If not, you may skip the remaining questions in this section.}
Yes, the paper abstracts were automatically segmented into sentences using the pySBD sentence tokenizer.

\bigskip
\noindent
22. \textbf{Was the ``raw'' data saved in addition to the preprocessed/cleaned/labeled data (for example, to support unanticipated future uses)? If so, please provide a link or other access point to the ``raw'' data.} The sentences sampled from abstracts can be accessed in the data, but we do not provide the full abstract texts. Users can use Semantic Scholar API\footnote{\url{https://www.semanticscholar.org/product/api}} with the paper IDs to download the corresponding full abstracts and use the open-source sentence segmenter to segment the abstracts to reproduce the preprocessed sentences.

\bigskip
\noindent
23. \textbf{Is the software that was used to preprocess/clean/label the data available? If so, please provide a link or other access point.} Yes, the sentence segmenter we use is available on Github: \url{https://github.com/nipunsadvilkar/pySBD}. We set the language to English. The Prodigy software is also available to download: \url{https://prodi.gy/}.

\bigskip
\noindent
24. \textbf{Has the dataset been used for any tasks already? If so, please provide a description.} The dataset was used by the authors to build value classifiers and use these classifiers to analyze  how different research communities in computer science express their contributions. 

\bigskip
\noindent
25. \textbf{Is there a repository that links to any or all papers or systems that use the dataset? If so, please provide a link or other access point.} This paper is currently the only one that uses the dataset. 

\bigskip
\noindent
26. \textbf{What (other) tasks could the dataset be used for?} The dataset could be used for several additional tasks: studying the evolution of writing styles across scientific communities, developing automated tools to suggest appropriate venues for paper submissions based on their value expression, investigating the relationship between value expression and paper impact, and building writing assistance tools for academic writing.

\bigskip
\noindent
27. \textbf{Is there anything about the composition of the dataset or the way it was collected and preprocessed/cleaned/labeled that might impact future uses? For example, is there anything that a dataset consumer might need to know to avoid uses that could result in unfair treatment of individuals or groups (for example, stereotyping, quality of service issues) or other risks or harms (for example, legal risks, financial harms)? If so, please provide a description. Is there anything a dataset consumer could do to mitigate these risks or harms?} Since the dataset samples from established academic venues in computer science, it may underrepresent work from emerging venues, less prominent institutions, or non-English publications, which could lead to biases if the models trained on this data are used to evaluate or suggest research values for papers from these underrepresented sources. In addition, since 10 popular values have been annotated in this dataset, models trained on this dataset might not capture the full spectrum of research values. 

\bigskip
\noindent
28. \textbf{Are there tasks for which the dataset should not be used? If so, please provide a description.} The dataset should not be used for automated paper acceptance decisions or evaluating individual researchers' work quality, as it only captures a limited aspect of research contribution (value expression in abstracts) and comes from a specific subset of venues. In addition, since the annotations are based on abstract sentences only, the dataset should not be used to make conclusions about how research values are expressed in full papers or other sections of academic writing.

\bigskip
\noindent
29. \textbf{How will the dataset be distributed (for example, tarball on website, API, GitHub)? Does the dataset have a digital object identifier (DOI)?} The dataset is published at \url{https://github.com/hjian42/CSValues}. No, it does not have a DOI.

\bigskip
\noindent
30. \textbf{When will the dataset be distributed?} The dataset will be distributed on February 24 2025.

\bigskip
\noindent
31. \textbf{Will the dataset be distributed under a copyright or other intellectual property (IP) license, and/or under applicable terms of use (ToU)? If so, please describe this license and/ or ToU, and provide a link or other access point to, or otherwise reproduce, any relevant licensing terms or ToU, as well as any fees associated with these restrictions.} The dataset will be distributed under an Open Data Commons Attribution License (ODC-By) v1.0.

\bigskip
\noindent
32. \textbf{Have any third parties imposed IP-based or other restrictions on the data associated with the instances? If so, please describe these restrictions, and provide a link or other access point to, or otherwise reproduce, any relevant licensing terms, as well as any fees associated with these restrictions.} No, there are no IP-based or other restrictions imposed by third parties on the data, as the dataset is composed of publicly available sentences from research paper abstracts.

\bigskip
\noindent
33. \textbf{Do any export controls or other regulatory restrictions apply to the dataset or to individual instances? If so, please describe these restrictions, and provide a link or other access point to, or otherwise reproduce, any supporting documentation.} No export controls or other regulatory restrictions apply to this dataset, as it consists of publicly available sentences from research paper abstracts.

\bigskip
\noindent
34. \textbf{Who will be supporting/hosting/maintaining the dataset?} The first author of the paper will be responsible for supporting and maintaining the dataset.

\bigskip
\noindent
35. \textbf{How can the owner/curator/manager of the dataset be contacted (for example, email address)?} The dataset owner can be contacted via email at \url{hjian42@icloud.com}.

\bigskip
\noindent
36. \textbf{Is there an erratum? If so, please provide a link or other access point.} No.

\bigskip
\noindent
37. \textbf{Will the dataset be updated (for example, to correct labeling errors, add new instances, delete instances)? If so, please describe how often, by whom, and how updates will be communicated to dataset consumers (for example, mailing list, GitHub)?} No, the dataset will not be updated after its initial release.

\bigskip
\noindent
38. \textbf{If the dataset relates to people, are there applicable limits on the retention of the data associated with the instances (for example, were the individuals in question told that their data would be retained for a fixed period of time and then deleted)? If so, please describe these limits and explain how they will be enforced.} This question is not applicable since the dataset consists of publicly available research paper abstracts and does not contain personal data. 

\bigskip
\noindent
39. \textbf{Will older versions of the dataset continue to be supported/hosted/maintained? If so, please describe how. If not, please describe how its obsolescence will be communicated to dataset consumers.} The question of supporting older versions is not applicable since this will be the first release of the dataset.

\bigskip
\noindent
40. \textbf{If others want to extend/augment/build on/contribute to the dataset, is there a mechanism for them to do so? If so, please provide a description. Will these contributions be validated/verified? If so, please describe how. If not, why not? Is there a process for communicating/distributing these contributions to dataset consumers? If so, please provide a description.} Yes, others can contribute to the project by submitting issues or pull requests through GitHub.

\end{document}